\documentclass[journal,onecolumn]{IEEEtran}
\usepackage{cite}

\ifCLASSINFOpdf
   \usepackage[pdftex]{graphicx}
\else
\fi

\usepackage{amsmath,amssymb} 
\usepackage{bbm}
\usepackage{float}
\usepackage{xcolor}
\usepackage{comment}
\newcommand{\todo}[1]{{\textcolor{black}{#1}}}

\begin{document}
%
\title{Scene text removal via cascaded text stroke detection and erasing}
%
%
%

\author{\IEEEauthorblockN{Xuewei BIAN${}^{1}$, Chaoqun WANG${}^2$, Weize QUAN${}^1$$^\ast$\thanks{$^\ast$Corresponding author. E-mail: qweizework@gmail.com}, Juntao YE${}^1$,  Xiaopeng ZHANG${}^1$, and Dong-Ming YAN${}^1$}
\IEEEauthorblockA{\textit{${}^1$NLPR, Institute of Automation, Chinese Academy of Sciences, Beijing {\rm 100190}, China} \\
\textit{${}^2$School of Artificial Intelligence, University of Chinese Academy of Sciences, Beijing {\rm 100049}, China}\\
}
}

\maketitle

\begin{abstract}
Recent learning-based approaches show promising performance improvement for scene text removal task. However, these methods usually leave some remnants of text and obtain visually unpleasant results. In this work, we propose a novel ``end-to-end'' framework based on accurate text stroke detection. Specifically, we decouple the text removal problem into text stroke detection and stroke removal. We design a text stroke detection network and a text removal generation network to solve these two sub-problems separately. Then, we combine these two networks as a processing unit, and cascade this unit to obtain the final model for text removal. Experimental results demonstrate that the proposed method significantly outperforms the state-of-the-art approaches for locating and erasing scene text. Since current publicly available datasets are all synthetic and cannot properly measure the performance of different methods, we therefore construct a new real-world dataset, which will be released to facilitate the relevant research.
\end{abstract}


%
\IEEEpeerreviewmaketitle

\section{Introduction}

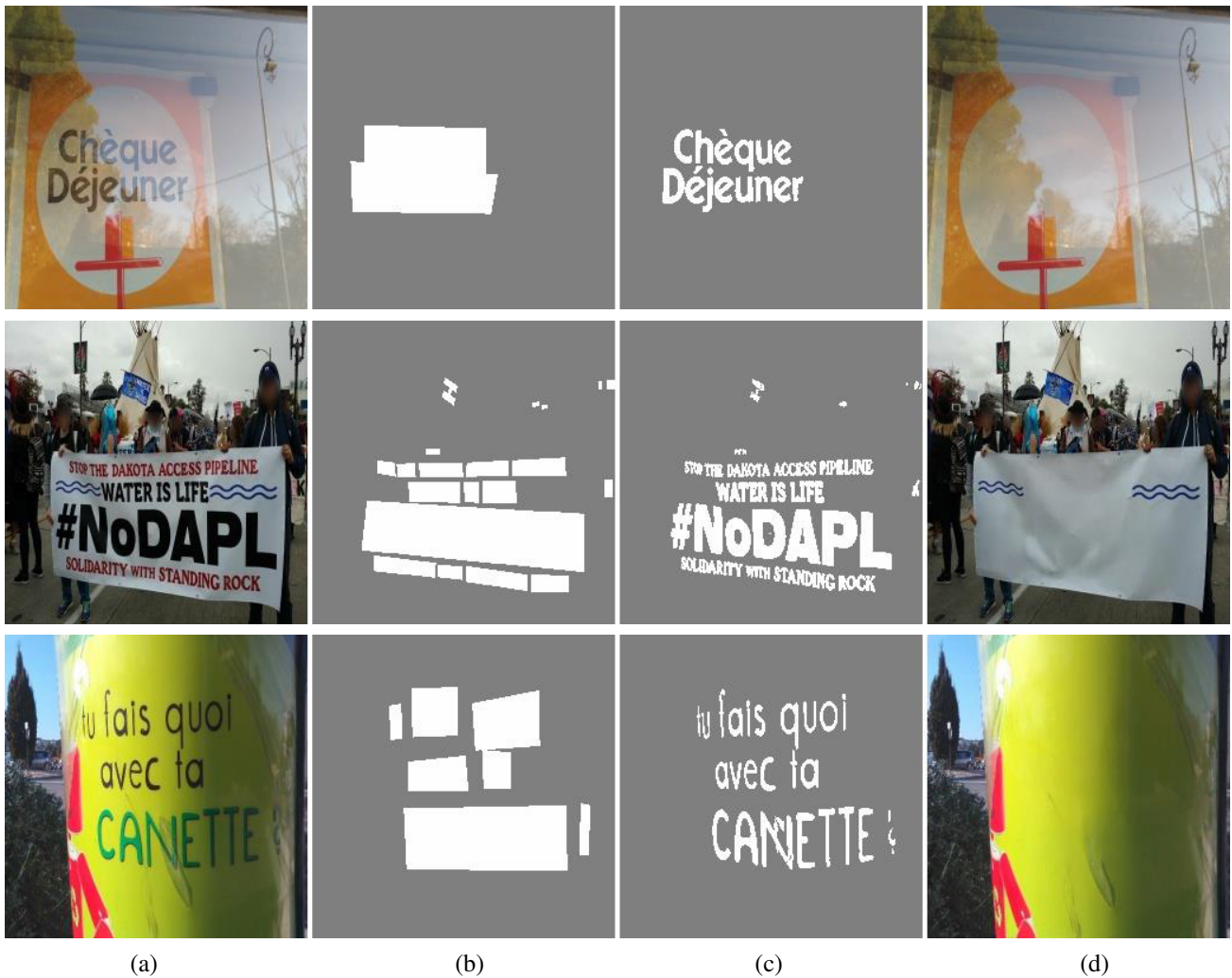
\begin{figure}[t]
\centering
\includegraphics[width=0.95\linewidth]{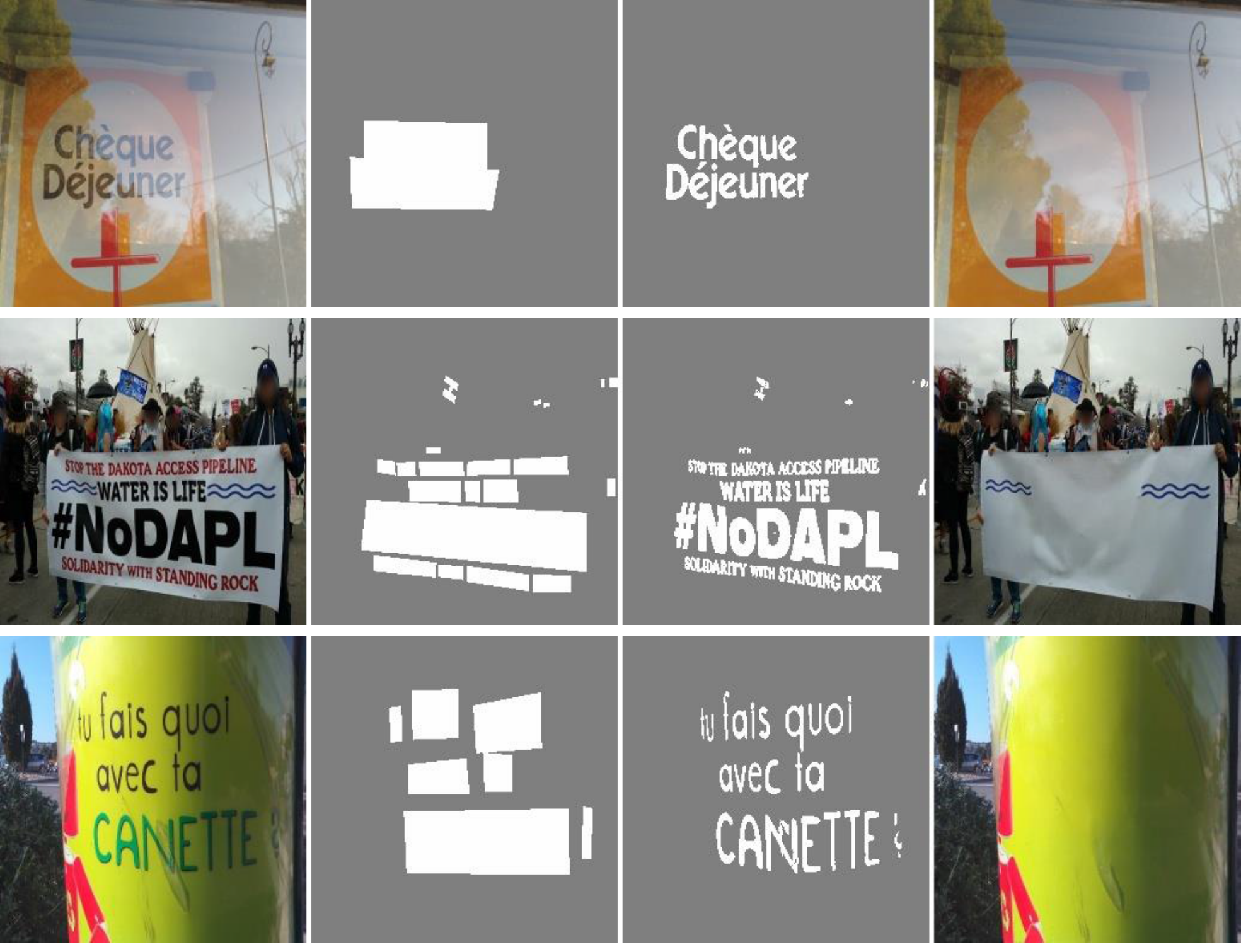}
\put(-440,-10){{(a)}}
\put(-310,-10){{(b)}}
\put(-190,-10){{(c)}}
\put(-60,-10){{(d)}}
\caption{Sampled results of the proposed scene text removal method. From left to right: (a) input image, (b) the region mask, (c) text stroke mask obtained by our TSDNet, and (d) the final result. Each row shows a representative result of commonly appeared scenarios.}
\label{fig:example1}
\end{figure}

Scene text is an important information carrier, and often appears in various scenarios. The problem of scene text removal can be stated as follows: given an image with appropriate amount of text [\textit{e.g.}, Fig.~\ref{fig:example1}(a)], the goal is to remove the text in this image [\textit{e.g.}, Fig.~\ref{fig:example1}(d)]. This task has many applications in our daily life, such as personal private information protection (hiding telephone numbers or home address from public photos), text translation (removing the original text and pasting new translated results), and so on.

Several approaches have been proposed to erase graphical text (\textit{e.g.}, subtitles) from color images~\cite{Khodadadi_2012_text,Modha_2012_image,Wagh_2015_text}. For the challenging scenario of scene text removal, which usually has complex background and text with various fonts and sizes, etc, however, these methods often produce results with visual artifacts. Inspired by the notable success of deep learning in image transformation~\cite{Johnson_2016_perceptual,Isola_2017_image,zhu_unpair_2017}, recent works have introduced deep-learning-based approaches to solve this problem and have achieved promising results~\cite{Nakamura_2017_scene,Zhang_2019_ensnet,Tursun_2019_mtrnet}. The learning-based methods can be roughly classified into two main categories, \textit{i.e.}, text removal without/with using mask. The former simply takes the given image as input and removes all the texts from the whole input image. This kind of methods often left noticeable remnants of text or distort non-text area incorrectly, and cannot remove text locally. The latter usually uses a region mask, \textit{i.e.}, a rectangle or polygon mask roughly indicating the text region [\textit{e.g.}, Fig.~\ref{fig:example1}(b)], as additional input to facilitate the text removal.

Recent MTRNet~\cite{Tursun_2019_mtrnet} achieved noticeable improvement compared to prior works for scene text removal, by focusing on text regions via auxiliary/binary mask. In fact, their pipeline is similar to the general image inpainting tasks~\cite{Iizuka_2017_global,Yu_2018_CVPR}. However, there is an apparent difference between them: for the text removal, the pixel values of original input image in the regions indicated by auxiliary mask (\textit{i.e.}, text regions) are known; whereas the corresponding values are unknown (corrupted) for the general image inpainting, \textit{i.e.}, the so-called missing regions. Generally speaking, when the regions to be processed (indicated by mask) are larger, it becomes harder to fill or remove the corresponding regions not only for the image inpainting, but also for the text removal. In addition, for the scene text removal problem itself, there is no need to remove regions not covered by text strokes like MTRNet. In other words, the mask used by MTRNet covers some unnecessary/redundant regions (\textit{i.e.}, non-stroke areas), especially when text strokes are scattered sparsely. It is obvious that if we can extract the exact text stroke, which means that we can preserve original contents of input image as much as possible, and then we could achieve better result. However, such precise areas are difficult to obtain, to best of our knowledge, there is no related research to focus on distinguishing text strokes from non-stroke area in the pixel-wise level.

In this paper, we propose a novel ``end-to-end'' framework based on \emph{generative adversarial network} (GAN) to address this problem. The key idea of our approach is first to extract text strokes as accurately as possible, and then improve the text removal process. These two processes can be further enhanced via a simple cascade. In addition, current public datasets for scene text removal are all synthetic, which to some extent affect the generalization ability of trained models. To facilitate this research and be close to real-world setting, we construct a new dataset with high quality. The main contributions of our work include:
\begin{itemize}
\item We design a text stroke detection network (TSDNet), which can effectively distinguish text strokes from non-text area.
\item We propose a text removal generation network and combine it with TSDNet to construct a processing unit, which is cascaded to obtain our final network. Our method demonstrates the superior performance.
\item We propose a weighted-patch-based discriminator to pay more attention to the text area of given images, making it easier for the generator to generate more realistic images.
\item We construct a high-quality real-world dataset for the scene text removal task, and this dataset can be used to benchmark related text removal methods. It can also be used in other related tasks.
\end{itemize}

The remainder of this paper is organized as follows. Section~\ref{sec:related_work} reviews relevant existing work. Section~\ref{sec:proposed_method} introduces the motivation and network details of our method. Section~\ref{sec:experiments} presents the performance evaluations for our method and detailed comparisons with existing methods. Section~\ref{sec:conclusion} draws the conclusions and discusses the future working directions.

\section{Related work}
\label{sec:related_work}

\subsection{Scene text detection}
Scene text detection is a fundamental step of scene understanding and is widely studied in the field of computer vision~\cite{ye_text_2015}. With the aid of deep learning, the performance of scene text detection framework has been significantly improved and surpassed traditional methods by large margins. Shi et al.~\cite{shi2017detect} decompose text into two locally detectable elements of segments and links, which are simultaneously detected by a fully-convolutional network. Liu et al.~\cite{liu_curved_2019} collect a curved text dataset called CTW1500 to facilitate the curved text detection task, and propose a method with the intergration of transverse and longitudinal sequence connection. Chen et al.~\cite{chen2019irregular} propose the concept of weighted text border and introduce attention module to boost the detection performance. To obtain better detection performance, multi-scale pyramid input is widely used with the consumption of much more running time. He et al.~\cite{he_realtime_2020} achieve remarkable speedup via a novel two-stage framework including a scale-based region proposal network and a fully convolutional network. CRAFT~\cite{baek_2019_craft} effectively detect arbitrary text area by exploring each character and affinity between characters. In this work, we adopt this method as the tool to measure the performance of scene text removal (more details in Section~\ref{subsec:metrics}).

\subsection{Text/non-text image classification}
Another relevant research is text/non-text image classification, which identifies whether a image block contains text or not. Zhang et al.~\cite{zhang_automatic_2015} first propose an effective method for text image discrimination, which is the suitable combination of maximally stable extremal region (MSER)~\cite{mser2004}, CNN, and bag of words (BoW)~\cite{bow1998}. Bai et al.~\cite{bai_text_2017} propose a mutli-scale spatial partition network to efficiently solve this task by predicting all image blocks simultaneously in a single forward propagation. Zhao et al.~\cite{zhao_fast_2019} investigate this task from two perspectives of the speed and the accuracy. They use a small and shallow CNN to accomplish high speed and then apply the knowledge distillation to improve its performance. Very recently, Gupta and Jalal ~\cite{gupta_text_2020} combine a text detector EAST~\cite{Zhou_2017_EAST} and classification subnetwork to achieve text/non-text image classification. Different from these previous works, our method mainly captures the exact position of text stroke, \textit{i.e.}, in the pixel-wise level, instead of image block/patch-wise level, to effectively facilitate the text removal network.

\subsection{Scene text removal}
Existing approaches of the scene text removal can be classified into two major categories: traditional non-learning methods and deep-learning-based methods.

Traditional approaches typically use color-histogram-based or threshold-based methods to extract text areas, and then propagate information from non-text regions to text regions depending on pixel/patch similarity~\cite{Khodadadi_2012_text,Modha_2012_image,Wagh_2015_text}. These methods are suitable for simple cases, \textit{e.g.}, clean and well focused text, whereas they have limited performance on complex scenarios, such as perspective distortion and complicated background, etc.

Recent learning-based approaches try to solve this problem with the powerful learning capacity of deep neural networks. Nakamura et al.~\cite{Nakamura_2017_scene} first propose a scene text erasing method (ST Eraser) based on convolutional neural network (CNN), and conducted text erasing patch by patch. This patch-based processing fails to localize text with complex shape and inevitably damaged the consistency and continuity of erased result. More recently, Zhang et al.~\cite{Zhang_2019_ensnet} design an end-to-end trainable framework (EnsNet) with a conditional GAN to remove text from natural images. Different from \cite{Nakamura_2017_scene} which erases text in an image patch by patch, EnsNet can erase the scene text on the whole image in an ``end-to-end" manner. For these two works, they do not use mask, and thus need to localize and remove text simultaneously. Such kind of methods often suffer from inaccurate text localization and incomplete text removal. To solve this, Tursun et al.~\cite{Tursun_2019_mtrnet} develop a mask-based text removal network (MTRNet). Auxiliary mask is used to provide information on where the text is, and enables MTRNet to focus on text removal better. The additional information provided by mask is the main reason why MTRNet outperforms previous studies. MTRNet also supports partial/local text removal by providing mask purposefully. All of these existing approaches often leave some text strokes unchanged or generate unpleasant contents because they cannot appropriately and exactly pay attention to the text strokes. Another shortcoming of current methods is that their training datasets are all synthetic, because the collection of real-world datasets are difficult and time-consuming.

In addition, a closely related problem to scene text removal is the general image inpainting, which aims at synthesizing plausible contents to fill missing/hole regions of the corrupted input images. Image inpainting has been extensively studied with the aid of deep learning methods. More recently, several GAN-based approaches are proposed for this purpose~\cite{Iizuka_2017_global,Ren_2019_ICCV,Yu_2018_freeform} and show strong ability of generating reasonable contents for missing regions. We also use the GAN framework in our approach for text removal.

\section{Proposed method}
\label{sec:proposed_method}

Region mask is commonly used in general image inpainting studies to indicate which regions should be filled. MTRNet~\cite{Tursun_2019_mtrnet} first introduces region mask into text removal task, and achieves good performance. Yet such direct introduction and application is inappropriate because it ignores the difference between scene text removal and general image inpainting. Region mask is suitable for image inpainting where it properly specifies the to-be-filled region. When directly applying in the text removal mask like MTRNet, unfortunately, it cannot reach pixel-level accuracy in distinguishing which regions are text strokes. Regarding the above inappropriateness, we believe that a more accurate text stroke mask can help to improve the performance of text removal methods. We therefore propose to decouple the text removal task into two sub-tasks: text stroke detection and stroke removal, and solve them separately. In the following, we explain the details of our network architectures and the training losses used in our network.

\subsection{Network architecture}

We design and implement a text stroke detection network, and then combine it with our proposed text removal generation network to construct a processing unit. The final network is obtained by cascading this unit and combining with a weighted-patch-based discriminator.

\subsubsection{Cascaded generator}
\begin{figure}[t]
\centering
\includegraphics[width=0.96\linewidth]{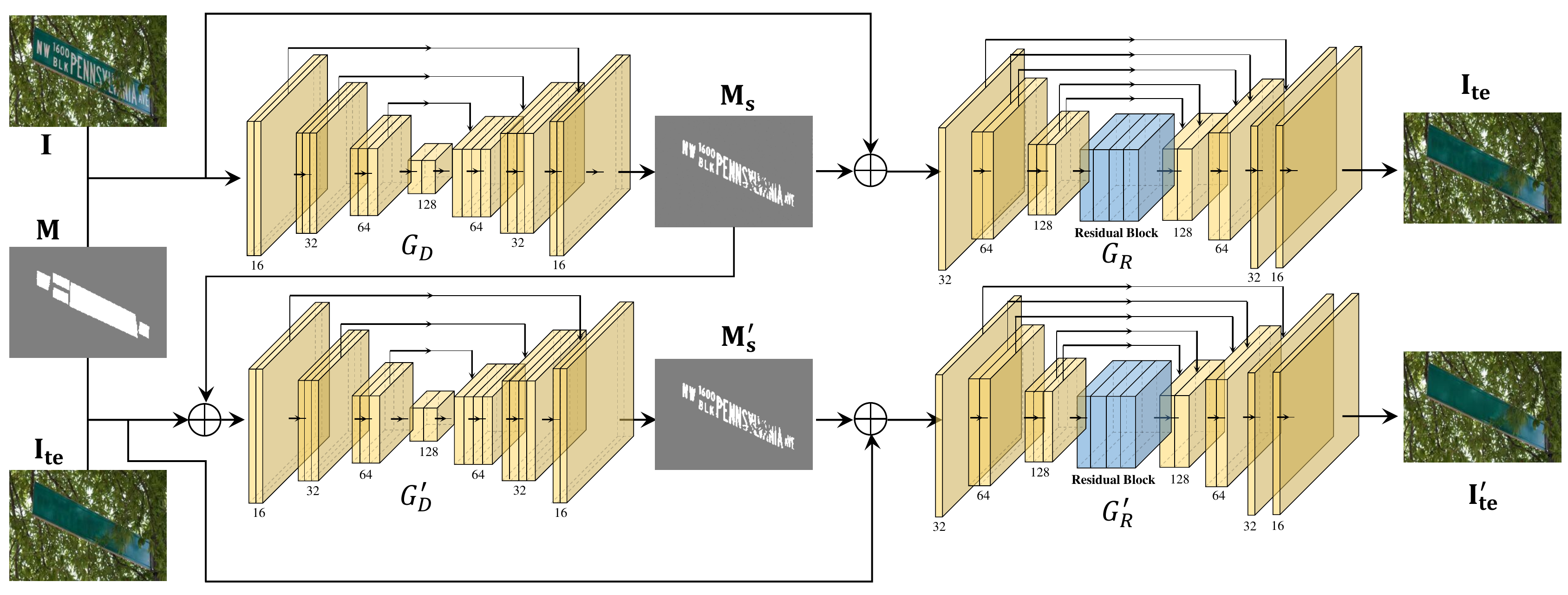}
\caption{The overall structure of the proposed generator, which consists of cascaded Text Stroke Detection and Text Removal Generation. $\oplus$ indicates the concatenation of image, region mask and stroke mask. The convolutional kernel size of the first layer of $G_R$ and $G_R'$ is $5 \times 5$, and the remaining kernel size is $3 \times 3$ in our proposed generator.}
\label{fig:generator}
\end{figure}

The proposed generator is designed for the following two purposes, \textit{i.e.}, 1) to detect text strokes in the input image accurately; 2) to inpaint the detected text strokes with proper content. To achieve the first goal, we construct a text stroke detection network (TSDNet). For the second goal, we propose a text removal generation network (TRGNet). The whole generator is obtained by cascading the group of TSDNet and TRGNet, as shown in Fig.~\ref{fig:generator}. Note that, the parameters in these four networks are not shared. Technically, both TSDNet and TRGNet employ a U-Net-like architecture~\cite{unet}, since by comparing with simple encoder-decoder framework, the U-Net architecture with skip connection helps to recover the structure and the texture details of unmasked area from input images, as well as to avoid over-smoothing and undesired artifacts to some extent.

The inputs of the TSDNet (noted as $G_D$), are a text image $\mathbf{I}$ and a binary mask $\mathbf{M}$ (indicating the text regions). The output is a float matrix $\mathbf{M_s}$ with the same size as $\mathbf{M}$, ranging from $0$ to $1$, in which larger value indicates higher confidence that corresponding position of image $\mathbf{I}$ is covered by text stroke.

\begin{equation}
    \mathbf{M_s} = G_D(\mathbf{I}, \mathbf{M}).
\end{equation}
The ground-truth of text stroke distribution is a binary mask $\mathbf{M_{gt}}$, in which $1$ means corresponding position of $\mathbf{I}$ is covered by text stroke. Different from $\mathbf{M}$ which only specifies the rough region of certain text, $\mathbf{M_{gt}}$ is a pixel-level annotation of text strokes. Practically, $\mathbf{M_{gt}}$ can be obtained by binarizing the difference between paired text image $\mathbf{I}$ and text-free image $\mathbf{I_{gt}}$ (see more details in Section~\ref{subsec:dataset}), and this stroke annotation is only used in the training stage as the supervised information to train TSDNet.

After obtaining the stroke mask $\mathbf{M_s}$, the TRGNet $G_R$ is then applied to erase text from the input image $\mathbf{I}$. $G_R$ takes three items as input, namely, text image $\mathbf{I}$, binary mask $\mathbf{M}$, and obtained stroke mask $\mathbf{M_s}$ from $G_D$, and outputs text-erased image $\mathbf{I_{te}}$, \textit{i.e.,}

\begin{equation}
    \mathbf{I_{te}} = G_R(\mathbf{I}, \mathbf{M}, \mathbf{M_s}).
\end{equation}
A TSDNet followed by a TRGNet (the first row of Fig.~\ref{fig:generator}) can already detect and erase text effectively, but the result images ($\mathbf{I_{te}}$) sometimes contain awkward artifacts and slight remnants of text. We observed that a simple cascade can eliminate such artifacts and bring visual improvement significantly. The designed architecture is as follows: the second TSDNet $G_D'$ takes $\mathbf{I_{te}}$, $\mathbf{M}$, and $\mathbf{M_s}$ as input, and outputs $\mathbf{M_s'}$. Then, the second TRGNet $G_R'$ takes $\mathbf{I_{te}}$, $\mathbf{M}$, and $\mathbf{M_s'}$ as input, and outputs the final text-erased result $\mathbf{I_{te}'}$. Through combining the previous outputs, $G_D'$ obtains more accurate text stroke distribution in an incremental manner, and thus $G_R'$ can reduce artifacts and inconsistency effectively.

Previous studies such as EnsNet and ST Eraser, which simply take an image/patch as input and try to erase text without any prior information, showed relatively limited performance. In this work, we use a binary mask (specifying the text region) as additional information to decrease the difficulty of detecting and erasing text at the same time, and design a TSDNet to provide more accurate instruction on which area should be removed. By doing so, we successfully decouple text removal into text stroke detection and stroke removal, and propose an effective solution and framework to solve these decoupled problems.

\subsubsection{Weighted-patch-based discriminator}
\begin{figure}[t]
\centering
\includegraphics[width=0.7\linewidth]{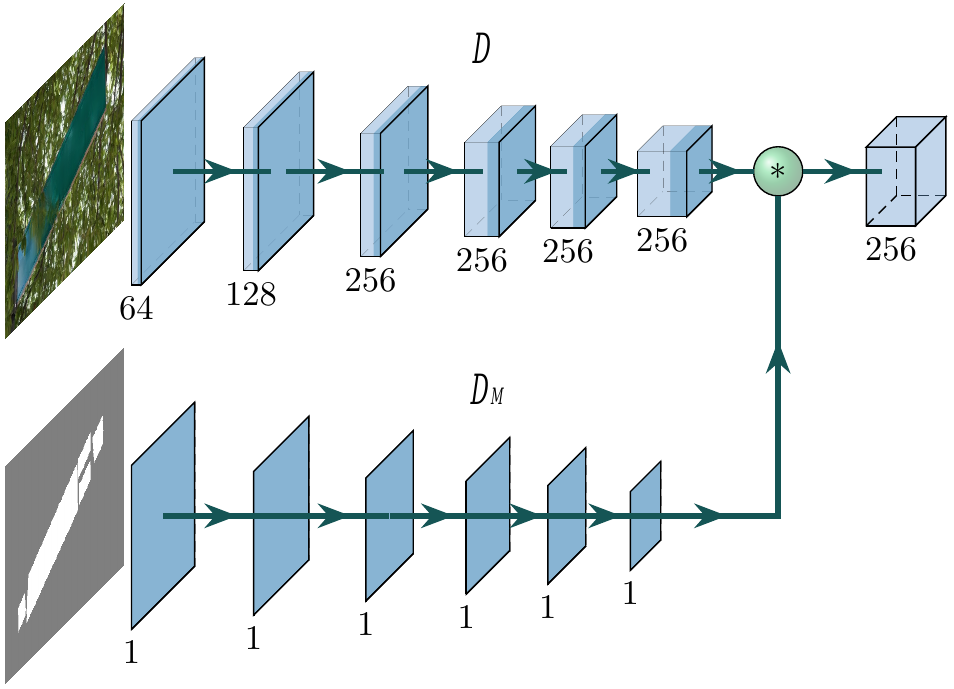}
\caption{Architecture of our proposed weighted-patch-based discriminator. * means element-wise multiplication between two branches with broadcasting. The convolutional kernel size is $5 \times 5$.}
\label{fig:disc}
\end{figure}

As text removal only needs to alter partial content of input image, thus a patch-based discriminator (see Fig.~\ref{fig:disc}) is more suitable to effectively concentrate on altered areas. In this work, we use the discriminator proposed in SN-PatchGAN~\cite{Yu_2018_freeform,Miyato-2018-spectral} to discriminate the text-erased image patch by patch. We further improve the original discriminator by attaching an additional convolutional branch $D_M$ to discriminator $D$ for assigning different weights to different patches according to mask $\mathbf{M}$. The $D_M$ has the same architecture as the $D$, but each layer only has one channel and weights in convolutional kernel are fixed to 1. By doing so, the patches covered by more text will be paid with more attention.

\subsection{Training loss}
In this subsection, we present our loss functions for the generator and discriminator. To verify that our proposed method is valid, we use relatively simple loss function when training our network.
For TSDNet $G_D$ and $G_D'$, we use simple $l_1$ loss,

\begin{equation} \label{equ:tsd}
    \mathcal{L}_{TSD} = \mathbf{E} \Bigl[ \| \mathbf{M_s} - \mathbf{M_{gt}} \|_{1} + \lambda_t \cdot \| \mathbf{M_s'} - \mathbf{M_{gt}} \|_{1} \Bigr],
\end{equation}
where $\lambda_t$ balances the ${l}_1$ loss of $G_D$ and $G_D'$. We set $\lambda_t=10$ in all our experiments, as most text strokes have been detected by $G_D$.

For scene text removal task, our main goal is to remove the text and preserve the non-text regions, therefore, more attention have to be paid to masked area (indicated by $\mathbf{M}$), especially detected stroke area (indicated by $\mathbf{M_s}$/$\mathbf{M_s'}$). More precisely, we define the corresponding weight matrix $\mathbf{M_w}$ and $\mathbf{M_w'}$ for $G_R$ and $G_R'$ as following, respectively:

\begin{equation}
\begin{split}
&\mathbf{M_w} = \mathbbm{1} + \lambda_m \cdot \mathbf{M} + \lambda_{s} \cdot \mathbf{M_s},  \\
    &\mathbf{M_w'} = \mathbbm{1} + \lambda_m \cdot \mathbf{M} + \lambda_{s} \cdot \mathbf{M_s'},
\end{split}
\end{equation}
where $\mathbbm{1}$ has same shape as $\mathbf{M}$ and its all elements are 1. Then, the total loss of $G_R$ and $G_R'$ is defined as

\begin{equation}
\begin{split}
\label{equ:loss}
    \mathcal{L}_{TRG} = &\mathbf{E} \Bigl[ \| \mathbf{I_{te}} \odot \mathbf{M_w} - \mathbf{I_{gt}} \odot \mathbf{M_w} \|_{1} \\
    &+ \lambda_r \cdot \| \mathbf{I_{te}'} \odot \mathbf{M_w'} - \mathbf{I_{gt}} \odot \mathbf{M_w'} \|_{1} \Bigr],
\end{split}
\end{equation}
where $\odot$ is the element-wise product operation, and $\lambda_r$ is the balance parameter. In all our experiments, we set $\lambda_m=5$, $\lambda_{s}=5$, and $\lambda_r=10$.

For the objective function of patch-based GAN, we use the hinge version of adversarial loss~\cite{Tran_2017_hier,Zhang_2019_self}. The corresponding loss function for generator and discriminator are respectively defined as

\begin{equation} \label{equ:gen}
    \mathcal{L}_G^{sn} = - \mathbf{E} \Bigl[ D_M(\mathbf{M}) \odot D(\mathbf{I_{te}'}) \Bigr],
\end{equation}

\begin{equation} \label{equ:dis}
\begin{split}
    \mathcal{L}_D^{sn} = &\mathbf{E} \Bigl[ ReLU(1 - D_M(\mathbf{M}) \odot D(\mathbf{I_{gt}}))\Bigr] + \\
    &\mathbf{E} \Bigl[ ReLU(1 + D_M(\mathbf{M}) \odot D(\mathbf{I_{te}'}))\Bigr].
\end{split}
\end{equation}
Note that, $\odot$ means element-wise product with broadcasting in terms of depth.

To summarize, the total loss for our cascaded generator is the summation of Eqn.~\ref{equ:tsd}, \ref{equ:loss} and \ref{equ:gen}:

\begin{equation}
    \mathcal{L}_G = \mathcal{L}_{TSD} + \mathcal{L}_{TRG} + \mathcal{L}_G^{sn}.
\end{equation}


Moreover, we observed that the perceptual loss~\cite{Johnson_2016_perceptual} and the style loss~\cite{Gatys_2016_image} have no noticeable improvement for our task. One reason is that scene text is usually located in relatively flat area. The total variation loss~\cite{Aly_2005_image} has no apparent effect on the erased result either, and thus is not used in our method.

\section{Experimental results}
\label{sec:experiments}
To evaluate our proposed method quantitatively and qualitatively, we compare our method with recent state-of-the-art text removal methods and general image inpainting methods, on synthetic dataset and our collected real-world dataset. Ablation study is also conducted to evaluate different components of our network.

\subsection{Dataset}
\label{subsec:dataset}
To train the deep model for text removal task, paired text image and text-free image are required. However, it is difficult to obtain such paired data for real-world scene images, this is why synthetic datasets are used for constructing text removal dataset in previous approaches. Currently, there are only two synthetic datasets, \textit{i.e.}, the Oxford synthetic scene text detection dataset~\cite{Gupta_2016_synthetic} and the SCUT synthetic text removal dataset~\cite{Zhang_2019_ensnet}. These two datasets adopted the same synthetic technology proposed by~\cite{Gupta_2016_synthetic}, and shared the same drawback, \textit{i.e.}, given a text-free image as background, a few or more images with synthesized text are obtained. For instance, the Oxford dataset synthesized 800,000 images using only 8,000 text-free images, which means that each 100 text images are synthesized using the same background image, leading to insufficiency of the diversity of background. Such kind of repetition would cause negative affect to the generalization ability of models.

In addition, synthetic data is only an approximation of real-world data. Current text synthesis technologies cannot generate realistic enough text images, which would restrain the text removal ability of models trained on synthetic data. When existing text removal methods are trained on synthetic dataset, and then tested on real-world data, we found that there often exists obvious text remnants and unsatisfactory artifacts, which is quantitatively analyzed in Section~\ref{subsec:comparison}. To this end, we propose to construct a real-world dataset for the text removal scenario.

\begin{figure}[t]
\centering
\includegraphics[width=0.8\linewidth]{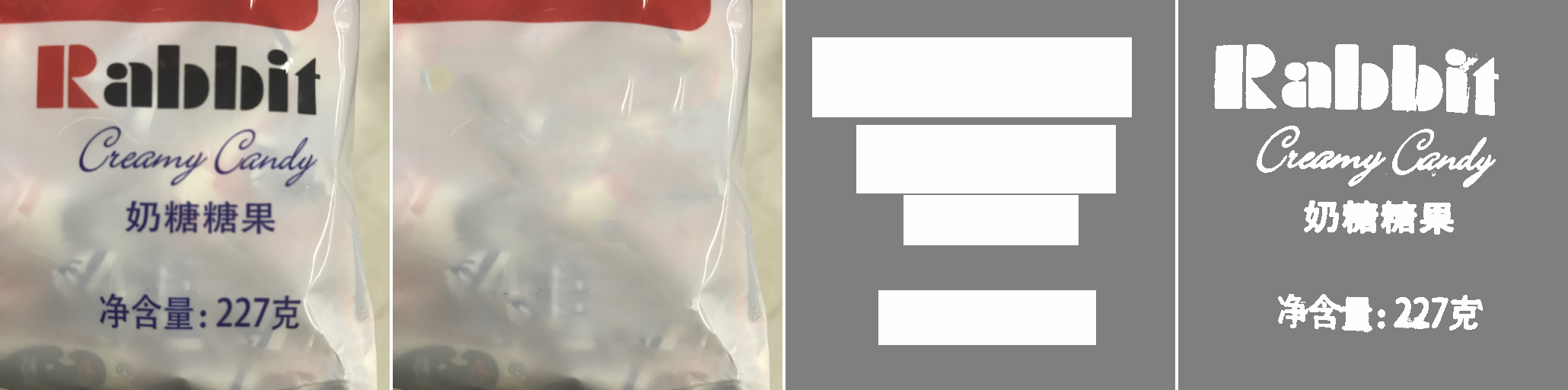}\\
\includegraphics[width=0.8\linewidth]{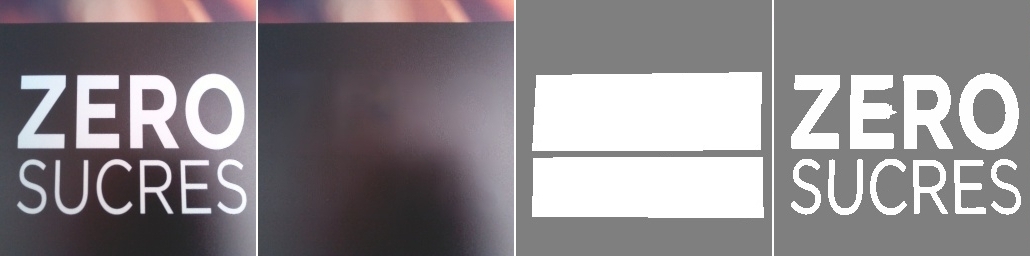}\\
\includegraphics[width=0.8\linewidth]{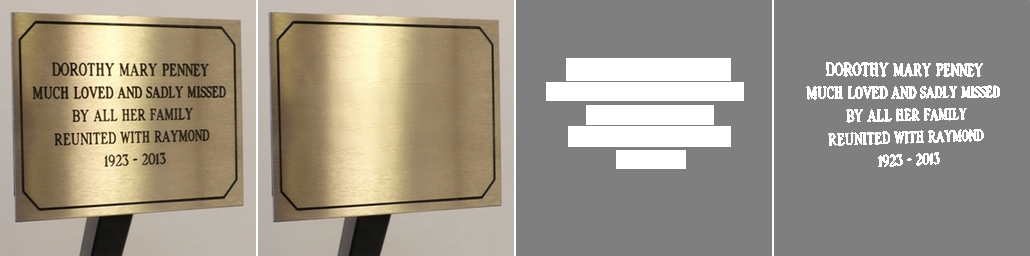}\\
\caption{Samples of our dataset. From left to right: images with text, text-free images, region masks, and text stroke masks.}
\label{fig:dataset_exam}
\end{figure}

\todo{To construct such dataset, we first collect 5,070 images with text from ICDAR2017 MLT dataset~\cite{Nayef_2017_mlt}, and 1,970 images captured from supermarkets and streets, etc. Then, post-processing are applied to obtain corresponding text-free images, region masks, and text stroke masks. We manually remove the text from these collected images and obtain text-free images as ground-truth using the inpainting tools in Photoshop$^\copyright$. The region masks are annotated by using VGG Image Annotator tool~\cite{dutta2019via}. For the ground-truth stroke mask, we first compute the difference between paired text images and text-free images, and then turn it into a binary image. To enrich the diversity of our dataset, we also use synthesis method and then manually select 4,000 images with high realism. In total, we obtain 11,040 images as training set (Train\_rw). Several samples of our dataset are shown in Fig.~\ref{fig:dataset_exam}. To construct testing set (Test\_rw), we additionally collect 1,080 real-world images and apply the same above post-processing to obtain the text-free images, region masks and text stroke masks.}

\todo{In this work, we mainly conduct the experiments and comparisons on our real-world dataset. In the meanwhile, we also conduct experiments on a public synthetic dataset, \textit{i.e.}, the Oxford dataset~\cite{Gupta_2016_synthetic}, which is much larger than the SCUT dataset~\cite{Zhang_2019_ensnet}. For the Oxford dataset, we randomly select $75\%$ of the whole dataset as training set (Train\_ox), and then randomly select 2,000 images from remaining set as testing set (Test\_ox). Note that, for these two datasets, \textit{i.e.}, our real-world (RW) dataset and Oxford dataset, there is no overlapping between training set and testing set.}

\subsection{Evaluation metrics}
\label{subsec:metrics}
We evaluate the performance from two different aspects: (1) can the method remove text from an image completely; (2) can text area be replaced with appropriate content. An accurate text detector is often used for the former evaluation metric. As for a text-erased image, the cleaner text is erased, the fewer text will be detected. In this work, we use the state-of-the-art text detector CRAFT~\cite{baek_2019_craft} and use DetEval protocol~\cite{Wolf_2006_image} for evaluation (\textit{i.e.}, recall, precision, and f-measure). For the second evaluation metric, we adopt general image inpainting metrics, and mainly use the following three indicators: 1) mean absolute error (MAE); 2) peak signal-to-noise ratio (PSNR); and 3) the structural similarity index (SSIM).

\subsection{Implementation details}
We implement our network using TensorFlow 1.13. The GPU version is TITAN RTX of NVIDIA\textsuperscript{\textregistered}~corporation. Input images are resized to $256 \times 256$. Adam optimizer~\cite{Kingma_2015_adam} with a minibatch size of 16 is used to train our network, and its $\beta_1$ and $\beta_2$ are set to 0.5 and 0.9 separately. Initial learning rate is set to 0.0001. The model is trained for 10 epochs on our dataset and 6 epochs on Oxford dataset.

\subsection{Dataset comparison}
\label{subsec:dataset comparison}

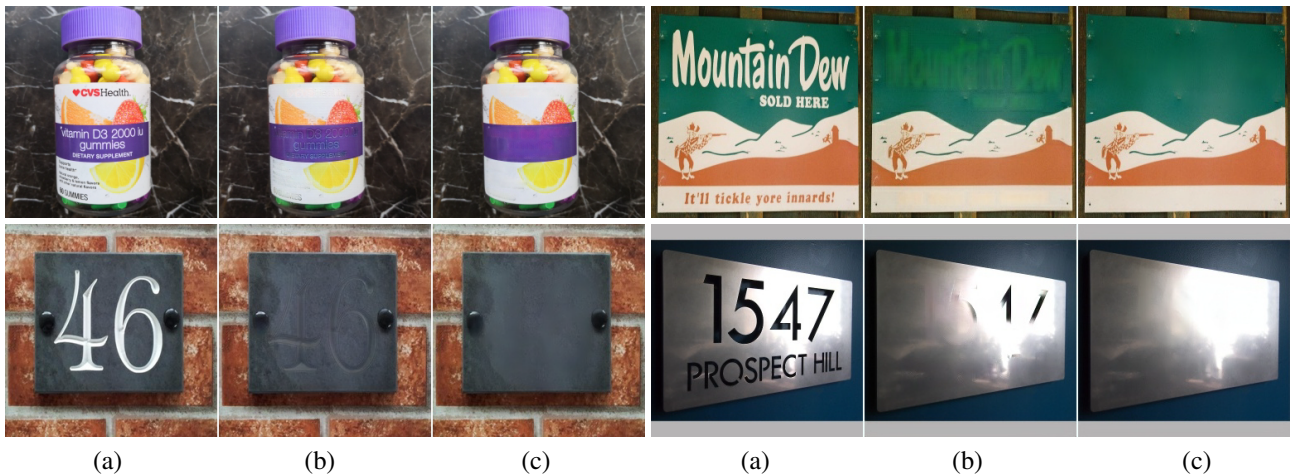
\begin{figure}
\centering
\includegraphics[width=0.95\linewidth]{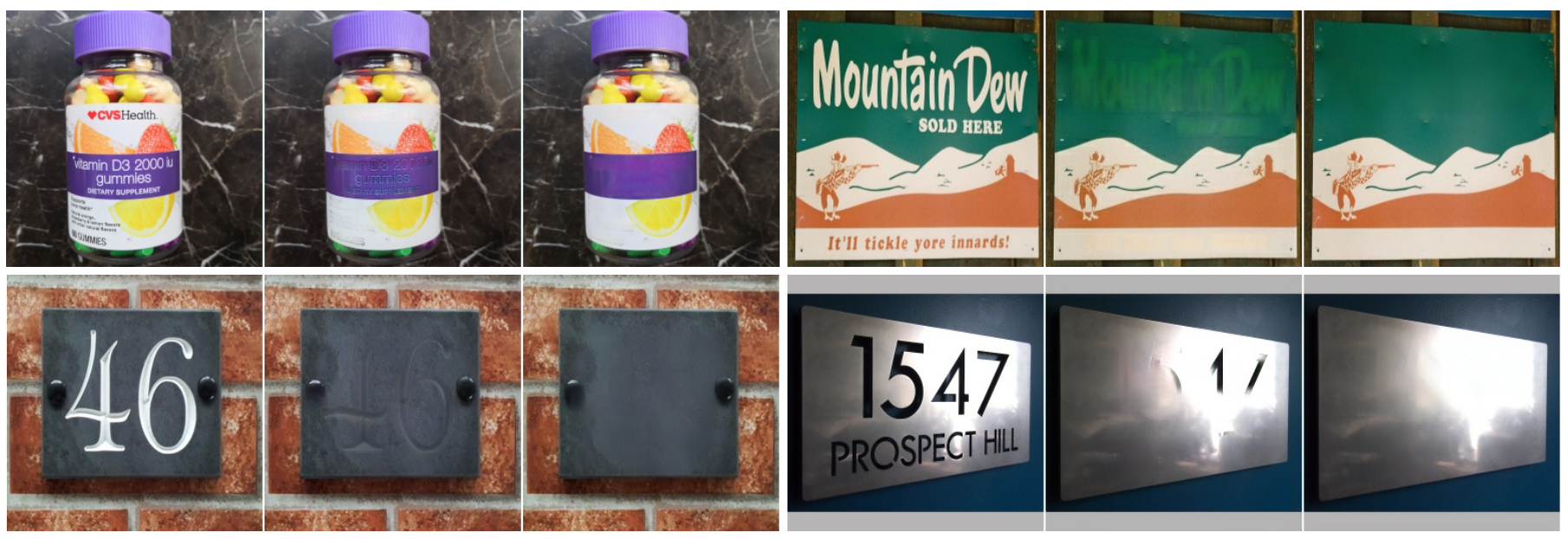}
\put(-455,-10){{(a)}}
\put(-375,-10){{(b)}}
\put(-293,-10){{(c)}}
\put(-210,-10){{(a)}}
\put(-130,-10){{(b)}}
\put(-43,-10){{(c)}}
\caption{Oxford dataset vs. Our dataset. We use different datasets to train the same method and compare the generalization ability of obtained models. The first row corresponds to the results of MTRNet and the second row corresponds to that of our method. Every three consecutive images (in row) form a group. For each group, from left to right: (a) the input image, (b) text removed result by model trained on the Oxford dataset, and (c) the text removed result by model trained on our dataset.}
\label{fig:dataset_comp}
\end{figure}

Fig.~\ref{fig:dataset_comp} shows the comparison between Oxford synthetic dataet and our real-world dataset using two different networks (MTRNet and our proposed network). Each group of images contains the input (left), the result of model trained on Oxford dataset (middle), and the result of model trained on our dataset (right). It is obvious that the text in right image of each group is better removed, especially for our method (the second row in Fig.~\ref{fig:dataset_comp}), which has no noticeable remnant and better preserves the original image details, such as lighting effect. This indicates that our dataset is more suitable for the scene text removal, even though the number of images in Oxford dataset is much larger than that of ours.

\subsection{Comparison with state-of-the-art methods}
\label{subsec:comparison}

We quantitatively and qualitatively compare our method with state-of-the-art text removal methods: ST Eraser~\cite{Nakamura_2017_scene}, EnsNet~\cite{Zhang_2019_ensnet}, and MTRNet~\cite{Tursun_2019_mtrnet}, as well as recent image inpainting method: GatedConv~\cite{Yu_2018_freeform}. We use the official implementation of EnsNet and GatedConv, and re-implemented ST Eraser and MTRNet.

Table~\ref{tab:comp} reports the quantitative comparison of the above five methods on the Oxford dataset and our dataset. We can find that our method is superior to other methods in MAE, PSNR,and SSIM by a large margin. When training on Train\_rw and testing on Test\_ox, our method achieves the best performance, and this phenomenon also exists when training on Train\_ox and testing on Test\_rw. This observation further indicates that our network has better generalization capability. For the cross-dataset validation, the results of training on Train\_ox and testing on Test\_ox are relatively similar with that of training on Train\_rw and testing on Test\_ox (see Table~\ref{tab:comp} column 9-14). However, the performance of training on Train\_rw and testing on Test\_rw is obviously better than that of training on Train\_ox and testing on Test\_rw (see Table~\ref{tab:comp} column 3-8), \textit{e.g.}, the PSNR and SSIM of EnsNet are improved by 7.37 (from 26.41 to 33.78) and 8.13$\%$ (from 87.30$\%$ to 95.43$\%$), respectively. These two results imply that our dataset is more suitable for this scene text removal task, especially for the real-world applications.

\begin{table*}[t]
\caption{Quantitative comparison of our method and state-of-the-art methods. All methods are trained and tested on the Oxford dataset and our dataset separately. For PSNR and SSIM (in $\%$), higher is better; For MAE, R (recall), P (precision), and F (f-measure), lower is better.}
\label{tab:comp}
\centering
\resizebox{1.\linewidth}{!}{
\begin{tabular}{cccccccccccccc}
\hline
 & Testing set & \multicolumn{6}{c}{Test\_rw} & \multicolumn{6}{c}{Test\_ox} \\ \hline
Training & Method & MAE & PSNR & SSIM & R($\%$) & P($\%$) & F($\%$) & MAE & PSNR & SSIM & R($\%$) & P($\%$) & F($\%$) \\ 
  set & Original image & - & - & -  & 43.25 & 40.68 & 41.93 & - & - & - & 48.24 & 67.93 & 56.42 \\ \cline{1-14}
 & ST Eraser(2017) & 2.52 & 27.20 & 91.13 & 6.23 & 20.55 & 9.56 & 2.67 & 27.94 & 90.24 & 14.42 & 49.27 & 22.31 \\ 
  & EnsNet(2019) & 1.22 & 33.78 & 95.43 & 1.94 & 20.18 & 3.53 & 1.89 & 31.37 & 93.03 & 7.25 & 49.84 & 12.65 \\ 
 Train\_rw& MTRNet(2019) & 1.62 & 34.31 & 96.34 & 0.55 & 17.14 & 1.06 & 2.31 & 31.81 & 92.36 & 0.49 & 37.27 & 0.96 \\ 
 & GatedConv(2019) & 1.42 & 34.82 & 96.10 & \textbf{0.04} & \textbf{1.49} & \textbf{0.08} & 2.50 & 30.42 & 89.52 & \textbf{0.01} & \textbf{6.25} & \textbf{0.03} \\ 
 & \textbf{Ours} & \textbf{0.75} & \textbf{39.44} & \textbf{97.56} & 0.35 & 10.23 & 0.68 & \textbf{1.63} & \textbf{34.40} & \textbf{93.97} & 0.05 & 15.38 & 0.10 \\ \hline
 & ST Eraser(2017) & 4.26 & 21.52 & 82.20 & 28.34 & 36.17 & 31.78 & 5.77 & 20.92 & 77.05 & 21.56 & 48.60 & 29.87 \\ 
  & EnsNet(2019) & 2.55 & 26.41 & 87.30 & 23.77 & 34.56 & 28.17 & 1.75 & 32.76 & 93.20 & 4.07 & 46.09 & 7.48 \\ 
 Train\_ox& MTRNet(2019) & 1.89 & 32.53 & 92.68 & 11.07 & 34.95 & 16.82 & 2.24 & 31.79 & 92.06 & 1.79 & 42.75 & 3.43 \\ 
 & GatedConv(2019) & 1.48 &  34.32 & 95.93 & \textbf{0.02} & \textbf{1.27} & \textbf{0.04} & 2.16 & 31.31 & 91.30 & 0.03 & 17.86 & 0.06 \\ 
 & \textbf{Ours} &\textbf{1.23} & \textbf{36.23} & \textbf{96.64} & 0.33 & 10.89 & 0.64 & \textbf{1.72} & \textbf{34.48} & \textbf{94.47} & \textbf{0.00} & \textbf{0.00} & \textbf{0.00} \\ \hline
\end{tabular}
}
\end{table*}

Our method has the best performance among four scene text removal methods when evaluated by recall, precision, and f-measure. These three metrics are often lowest for GatedConv, because GatedConv first fully removes the text area and then fill the so-called missing region. Such processing can avoid incomplete text removal, bringing lowest recall, precision and f-measure, but can also bring obvious over-smoothing and boundary inconsistency to inpainted area, as shown in the following qualitative comparison.

\begin{figure}[t!]
\centering
\includegraphics[width=0.95\linewidth]{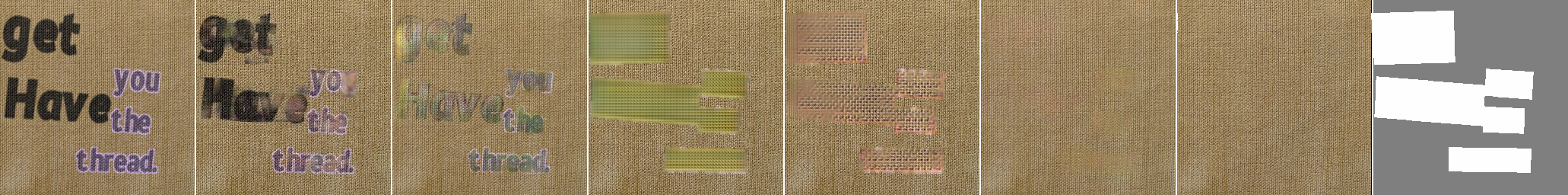}\\
\includegraphics[width=0.95\linewidth]{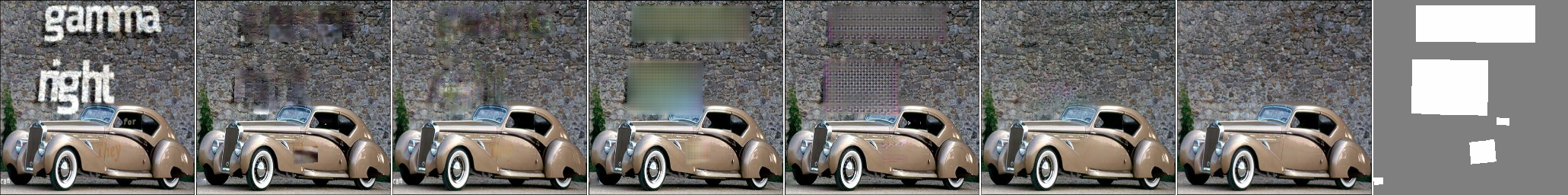}\\
\includegraphics[width=0.95\linewidth]{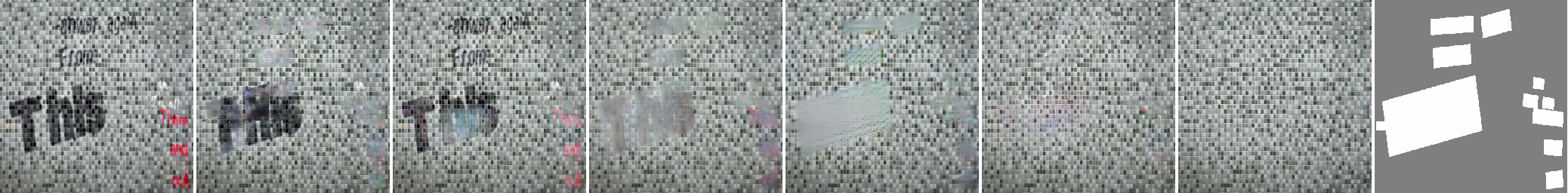}\\
\includegraphics[width=0.95\linewidth]{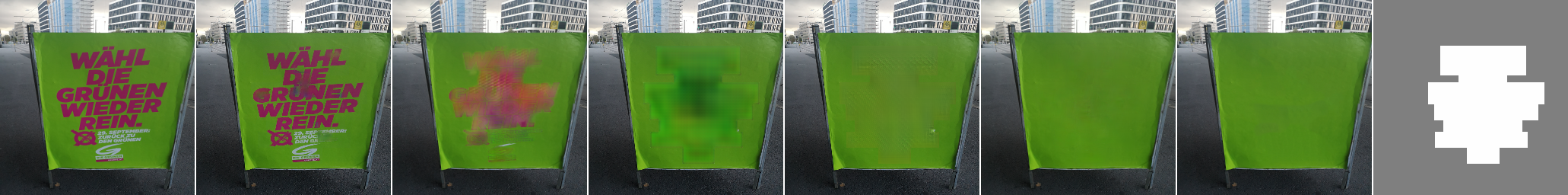}\\
\includegraphics[width=0.95\linewidth]{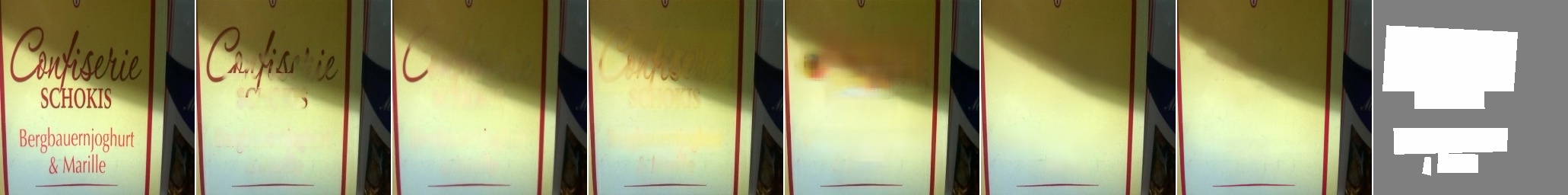}\\
\includegraphics[width=0.95\linewidth]{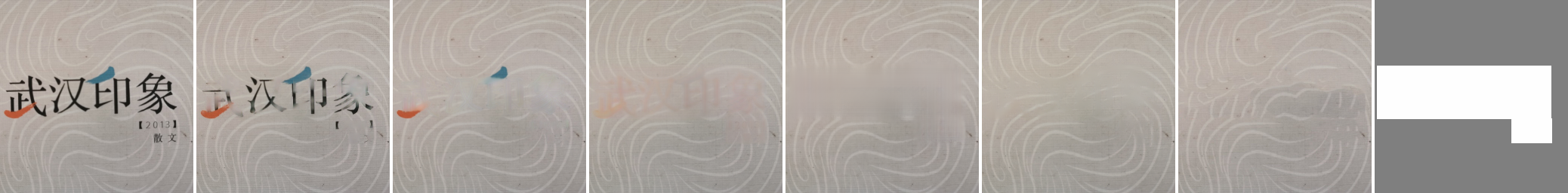}\\
\includegraphics[width=0.95\linewidth]{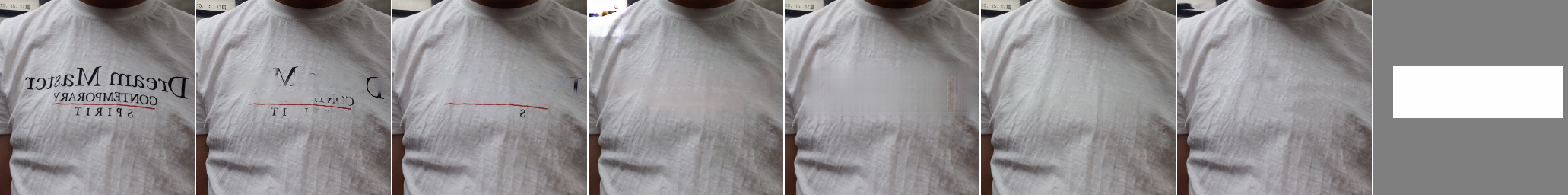}
\put(-470,-10){{Input}}
\put(-415,-10){{ST Eraser}}
\put(-350,-10){{EnsNet}}
\put(-294,-10){{MTRNet}}
\put(-225,-10){{Gated}}
\put(-160,-10){{Ours}}
\put(-96,-10){{GT}}
\put(-40,-10){{Mask}}
\caption{Qualitative comparison of all methods. The top three rows are synthetic images, and the rest are real-world images. From left to right: input image, ST Eraser, EnsNet, MTRNet, GatedConv, Our method, ground-truth, and input mask.}
\label{fig:visual_comp}
\end{figure}

\todo{Fig.~\ref{fig:visual_comp} shows the text-erased images of all five methods. Compared with other text removal methods, our method is more effective in erasing text and inpainting text area with proper content. In the first row of Fig.~\ref{fig:visual_comp}, our result preserves more consistent texture details with original non-text areas, whereas that of other methods has obvious text remnants or visual inconsistency. Comparing the results in the fifth row, our result has no text remnant and well maintains the original structure, \textit{i.e.}, the light transition. Furthermore, the text-erased images of our method show more reasonable texture details than that of GatedConv (comparing the fifth and sixth column in Fig.~\ref{fig:visual_comp}). The reason is that GatedConv first replaces the masked area with blank images, which will lose the useful detail information, thus resulting in the filling of texture itself becomes relatively more difficult. These results indicate that the reasonability of distinguishing text stroke area from non-text area in the masked region, which can guide the network to focus on text stroke area and preserve the useful information of original input image.}

\begin{figure}[htp]
\centering
\includegraphics[width=.85\linewidth]{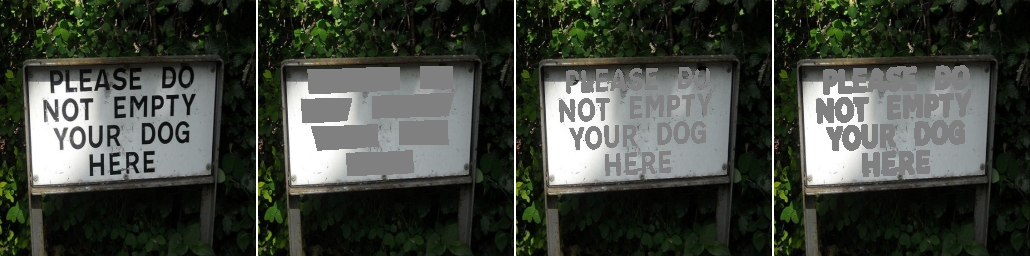}
\includegraphics[width=.85\linewidth]{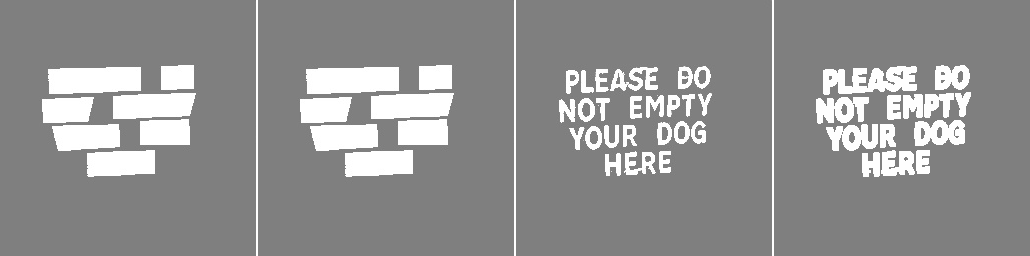}
\includegraphics[width=.85\linewidth]{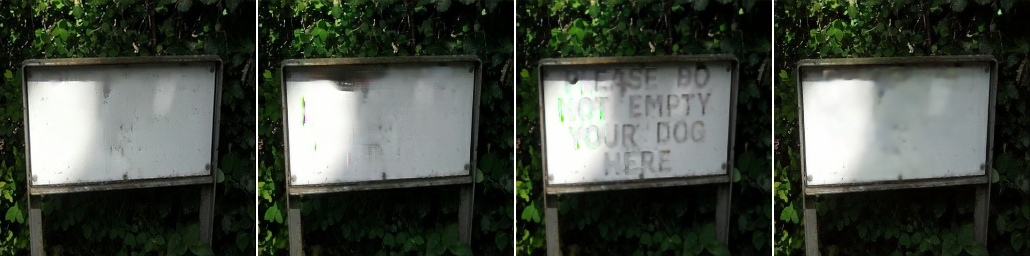}
\includegraphics[width=.85\linewidth]{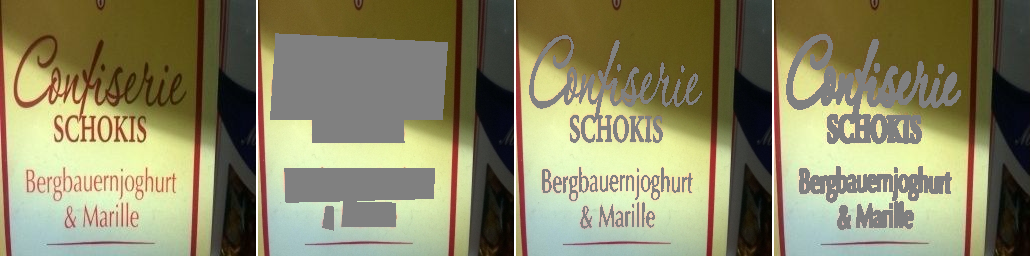}
\includegraphics[width=.85\linewidth]{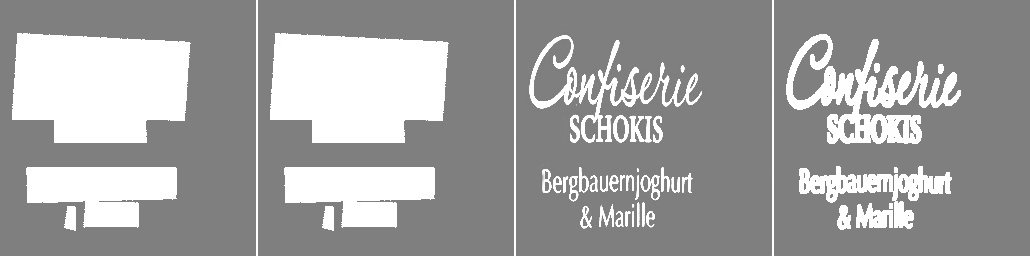}
\includegraphics[width=.85\linewidth]{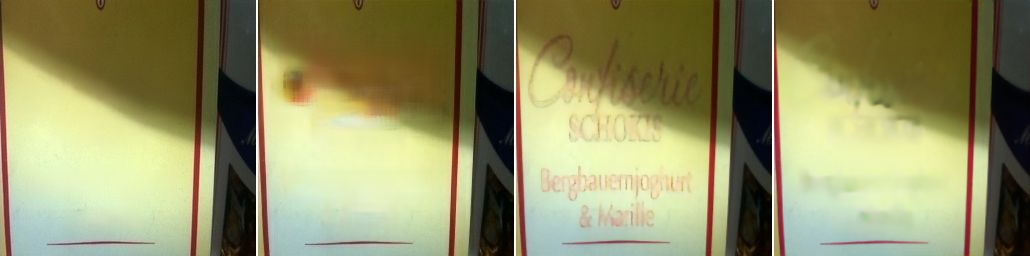}
\put(-390,-10){{(a)}}
\put(-280,-10){{(b)}}
\put(-167,-10){{(c)}}
\put(-60,-10){{(d)}}
\caption{Comparison of our method with a two-stage method. For each group (including three rows), the first and second rows are the inputs of network, and the third row is the final result. From left to right: (a) our method; (b) GatedConv with rectangle mask; (c) GatedConv with stroke mask, which is detected by our TSDNet; (d) GatedConv with dilated stroke mask.}
\label{fig:comp_gated}
\end{figure}

\todo{As analyzed earlier, using text region masks instead of text stroke masks is an important reason why existing scene text removal methods have limited performance. Inspired by this, our proposed novel and generic framework decouples the text removal problem into text stroke detection and stroke removal, and achieves superior performance. Following the pipeline of our ``end-to-end'' framework, a two-stage method, \textit{i.e.}, first extracting text strokes with a semantic segmentation algorithm and then filling these holes with image inpainting approach (\textit{e.g.}, GatedConv~\cite{Yu_2018_freeform}), may be possible to handle this task. Fig.~\ref{fig:comp_gated} compares the results of our method, GatedConv, and this two-stage method. In this experiment, we simply use the detected stroke mask via our TSDNet as the output of the first stage for the two-stage method. In the third and sixth row of Fig.~\ref{fig:comp_gated}, we find that the results of this two-stage method [(c) and (d)] both are visually unpleasant, whereas the results of our method [(a)] are consistently good. Noticeably, the final result of this two-stage method is sensitive to the output of the segmentation stage, \textit{i.e.}, the detected text stroke mask, when directly using the stroke mask provided by our TSDNet, there is plenty of text remnants [see Fig.~\ref{fig:comp_gated}(c)], and this phenomenon is improved by dilating the corresponding stroke mask [see Fig.~\ref{fig:comp_gated}(d)]. The possible reason is that GatedConv relies so much on the surroundings of the masked area that a tiny noise would effect the final text removal results noticeably. On the contrary, our ``end-to-end'' framework only takes the detected stroke mask of our TSDNet as the middle information, and thus is robust to stroke mask.}

\begin{figure}[t]
\centering
\includegraphics[width=0.9\linewidth]{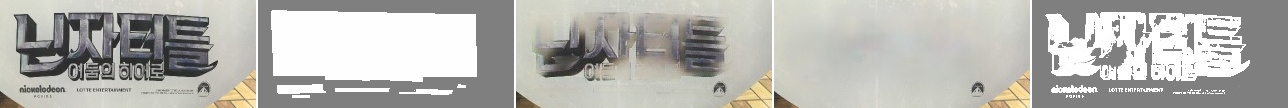}\\
\includegraphics[width=0.9\linewidth]{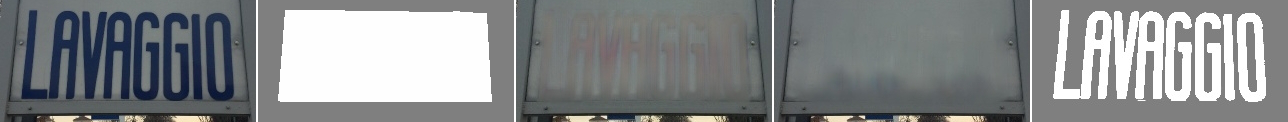}\\
\includegraphics[width=0.9\linewidth]{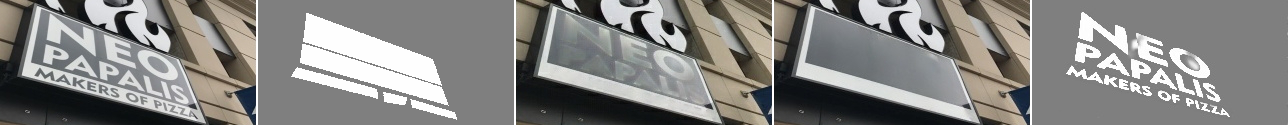}\\
\includegraphics[width=0.9\linewidth]{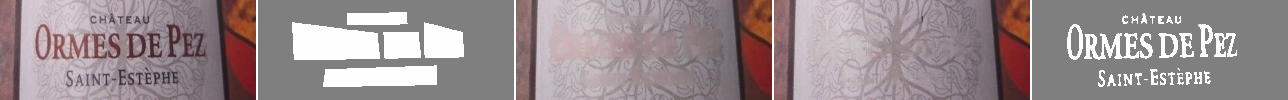}
\includegraphics[width=0.9\linewidth]{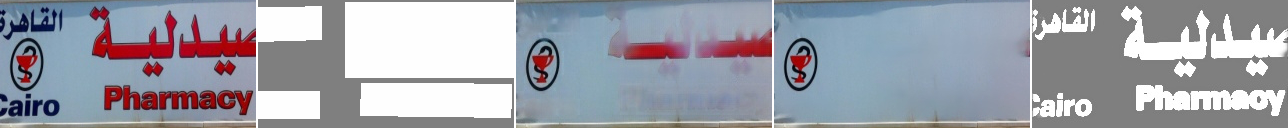}\\
\includegraphics[width=0.9\linewidth]{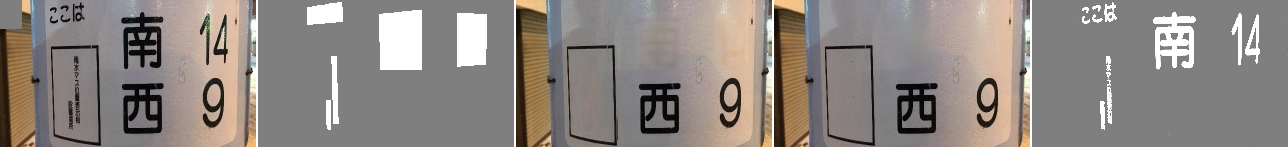}\\
\includegraphics[width=0.9\linewidth]{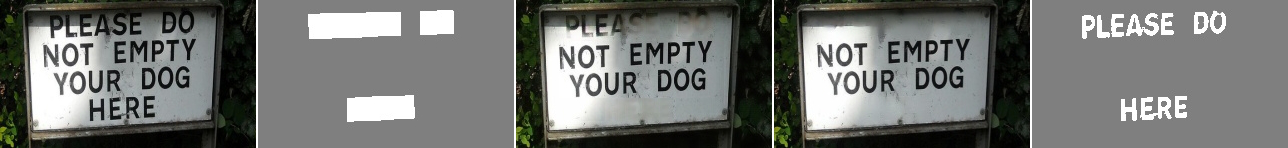}\\
\includegraphics[width=0.9\linewidth]{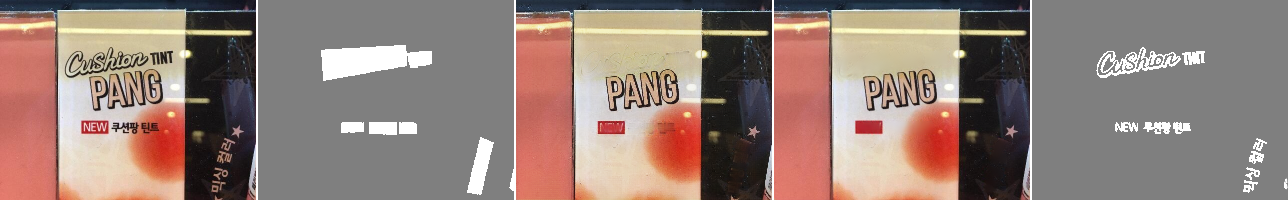}
\put(-430,-10){{Input}}
\put(-335,-10){{Mask}}
\put(-250,-10){{MTRNet}}
\put(-145,-10){{Ours}}
\put(-71,-10){{Stroke Mask}}
\caption{Results of multi-lingual text removal and selective text removal (the last three rows). From left to right: original image, input mask, MTRNet output, our output, and detected stroke mask of our method. Both models are trained on our dataset.}
\label{fig:example}
\end{figure}

\subsection{Multi-lingual text removal and selective text removal}

In this subsection, we also illustrate more results about multi-lingual text removal and selective text removal. We train MTRNet~\cite{Tursun_2019_mtrnet} and our method on our real-world dataset, and the corresponding results are reported in Fig.~\ref{fig:example}. Compared with MTRNet, our method can successfully remove the text in various languages. The reason is that our text stroke detection network focuses on the text, even learns the difference between various languages, and thus provides more useful information for the consecutive text removal generation network than the region mask.

In addition, our method can also accomplish the selective text removal. Given an auxiliary mask, where the desired removal text is indicating by a polygonal mask, our method can purposefully remove desired texts and does not affect the other text. Several sampled examples are shown in the last three rows of Fig.~\ref{fig:example}.

\begin{figure}[t]
\centering
\includegraphics[width=0.9\linewidth]{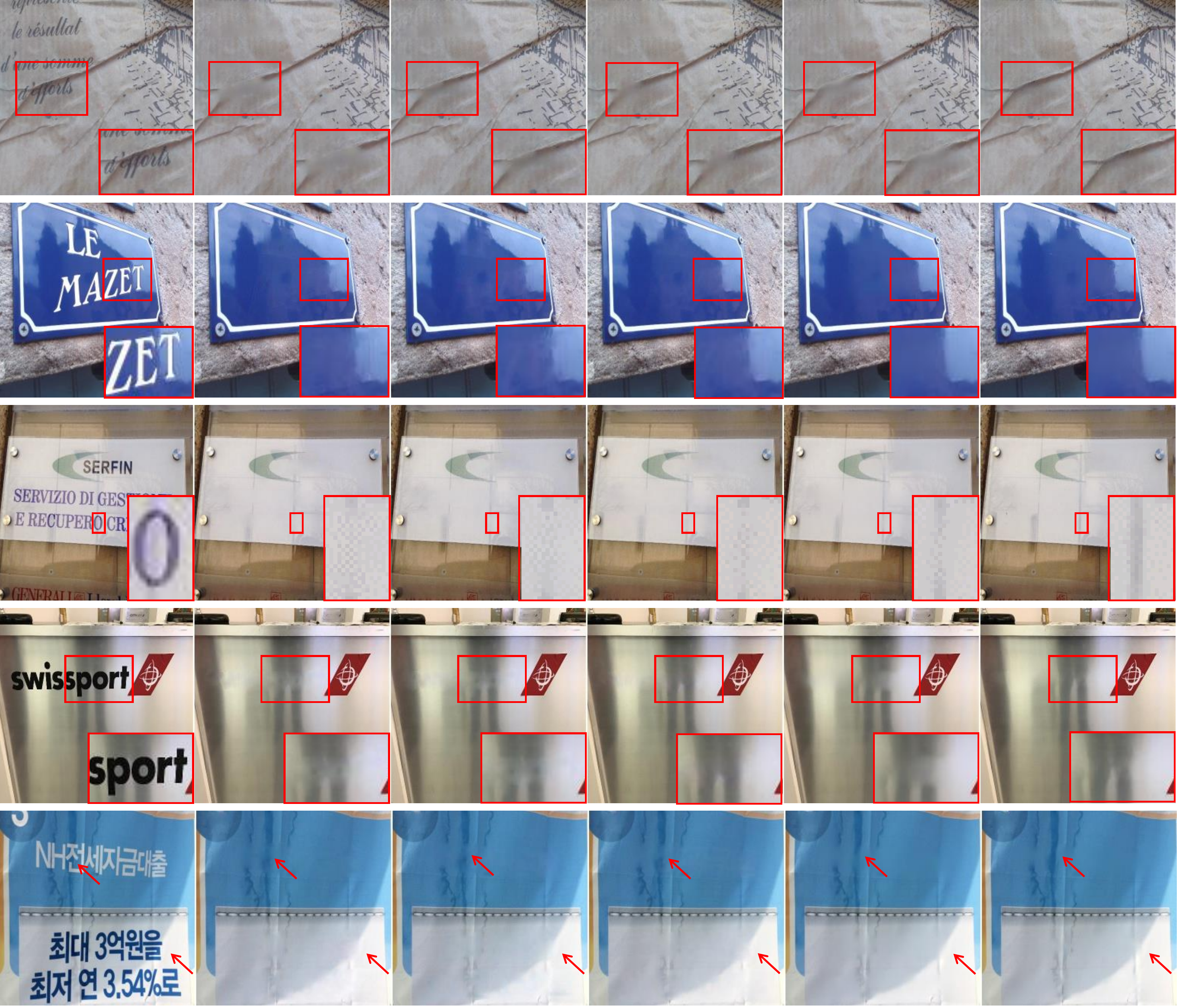}
\put(-440,-10){{Input}}
\put(-362,-10){{Baseline}}
\put(-277,-10){{WD}}
\put(-208,-10){{TSDNet}}
\put(-144,-10){{WD+TSDNet}}
\put(-56,-10){{Cascade}}
\caption{Qualitative results of ablation study. The last column gives the best result.}
\label{fig:ablation}
\end{figure}

\begin{table}[t]
\centering
\caption{Ablation study. Models are trained on Train\_rw and tested on Test\_rw. tMAE is the mean absolute error between detected stroke mask and ground-truth stroke mask.}
\label{tab:ablation}
\small
\begin{tabular}{ccccc}
\hline
Method & MAE & PSNR & SSIM($\%$) & tMAE($\%$) \\ \hline
Baseline & 1.59 & 35.00 & 95.42 & - \\ \hline
WD & 1.00 & 38.31 & 97.22 & - \\ \hline
TSDNet & 0.98 & 38.17 & 97.33 & 7.63 \\ \hline
WD + TSDNet & 0.92 & 38.47 & 97.48 & 4.85 \\ \hline
Cascade & \textbf{0.75} & \textbf{39.44} & \textbf{97.56} & \textbf{4.73} \\ \hline
\end{tabular}
\end{table}

\subsection{Ablation study}
Next, we study the effect of different components of our network. The corresponding results are reported in Table~\ref{tab:ablation} and Fig.~\ref{fig:ablation}.

\subsubsection{Baseline} For baseline model, we use a single TRGNet ${G}_R$ (shown in Fig.\ref{fig:generator}) as the generator, and use the discriminator proposed in SN-PatchGAN~\cite{Yu_2018_freeform} as the discriminator (\textit{i.e.}, $D$ in Fig.~\ref{fig:disc}). The inputs of TRGNet here are a text image $\mathbf{I}$ and a binary mask $\mathbf{M}$. Comparing Table~\ref{tab:comp} and~\ref{tab:ablation}, we find that such designed baseline model already shows similar performance with previous text removal methods, including EnsNet which does not use auxiliary mask and MTRNet which uses region mask. The visual results are also good as shown in the second column of Fig.~\ref{fig:ablation}.

\subsubsection{Weighted-patch-based discriminator (WD)} Original discriminator proposed in SN-PatchGAN treats all patches equally. In our work, masked regions indeed are the focus of our attention. We propose a weighted-patch-based discriminator, which can pay more attention to masked area via assigning higher weight. Comparing the rows of ``Baseline'' and ``WD'' in Table~\ref{tab:ablation}, it shows that our proposed discriminator can significantly improve the performance of the baseline model with the help of this weighted design.
For example, the first row in Fig.~\ref{fig:ablation} shows that our proposed weighted-patch-based discriminator can help to maintain the structure consistency of given image.

\subsubsection{Text stroke detection network (TSDNet)} A TSDNet is added into baseline model to prove the effectiveness of accurate text stroke extraction. As shown in Table~\ref{tab:ablation}, baseline model with TSDNet obtains much higher PSNR, SSIM, and much lower MAE (compare the rows of ``Baseline'' and ``TSDNet''). Our proposed TSDNet can effectively distinguish whether the given area is text stroke or not. And this useful information can help TRGNet to remove masked areas more purposefully. When combining the WD and TSDNet (see ``WD+TSDNet'' in Table~\ref{tab:ablation}), the performance can be further improved.
Comparing the results in the second row of Fig.~\ref{fig:ablation}, it can be seen that our TSDNet can help to completely remove text from image (see the character ``T'').

\subsubsection{Cascaded TSDNet and TRGNet (Cascade)} Cascading of TSDNet and TRGNet can help to fix minor mistakes and slight text remnants of the first unit of TSDNet and TRGNet, such as, completing partial-detected text stroke, removing text residual, and fixing visual artifacts. We also experiment three cascade, and the text-erased results are a little bit blurry. A possible reason is that part of high frequency information is lost during cascading.


\subsubsection{The effect of stroke detection}\todo{The text stroke detection is an important ingredient of our generic framework, here, we further discuss the effect of stroke detection performance on the final text removal. When inserting TSDNet into the Baseline, the performance is obviously improved, which validates our design of the TSDNet. Furthermore, the text stroke detection performance is enhanced via the cascaded design (tMAE of ``Cascade'' is significantly smaller than that of ``TSDNet'' in Table~\ref{tab:ablation}), in the meanwhile, the text removal performance is better. This further illustrates that the improvement of stroke detection can enhance the final text removal result.}

\section{Conclusion}
\label{sec:conclusion}
In this work, we proposed a novel GAN-based framework to solve scene text removal task via decoupling text stroke detection and stroke removal. We designed and implemented a text stroke detection network and a text removal generation network, and constructed the final model by cascading the group of above two networks. Quantitative and qualitative results illustrate the superior performance of our proposed network. Our study implies that it is beneficial to know the position of text strokes for the scene text removal problem. To the best of our knowledge, our study is the first to reveal the importance of accurate text strokes to text removal task. In the meanwhile, we also constructed a versatile real-world dataset, including text images, ground-truth text-free images, and auxiliary masks, which can be used to benchmark text removal methods. Moreover, our approach can be used for quickly constructing the large scale text-free image dataset from images with text, and pixel-wise text stroke annotations can be obtained as well (\textit{i.e.}, binarizing the difference between paired text image and text-free image). This kind of dataset will provide more and fine-grained supervised information to further improve the performance of scene text detection and recognition tasks.

Our method might generate implausible result if the text area is too large. We believe that a larger dataset with more diverse data can help to mitigate existing shortcomings. In the future, we plan to collect more real-world text images and construct a larger and richer dataset that can be used for both text removal task and other related research, \textit{e.g.}, realistic text synthesis. In this work, we use the text region masks in an ``off-line'' manner, considering not very perfect performance of the current automatic text detectors and the requirement of partial text removal applications. We would like to design a more ``complete'' framework combining automatic text detection, which supports the refinement of possible detection errors and the selection of specific text region with simple user guidance. It would also be interesting to study text removal problem in the semi-supervised manner. We plan to share our source code and real-world dataset to the research community.



%



\bibliographystyle{vancouver}
\bibliography{pr}

\begin{thebibliography}{10}

\bibitem{Khodadadi_2012_text}
Khodadadi M, Behrad A.
\newblock Text localization, extraction and inpainting in color images.
\newblock In: Proceedings of the Iranian Conference on Electrical Engineering;
  2012. p. 1035--1040.

\bibitem{Modha_2012_image}
Modha U, Dave P.
\newblock Image Inpainting - Automatic Detection and Removal of Text From
  Images.
\newblock International Journal of Engineering Research and Applications.
  2012;2(2):930--932.

\bibitem{Wagh_2015_text}
Wagh P~D, Patil D~R.
\newblock Text detection and removal from image using inpainting with
  smoothing.
\newblock In: Proceedings of the International Conference on Pervasive
  Computing; 2015. .

\bibitem{Johnson_2016_perceptual}
Johnson J, Alahi A, Fei-Fei L.
\newblock Perceptual losses for real-time style transfer and super-resolution.
\newblock In: European Conference on Computer Vision; 2016. p. 694--711.

\bibitem{Isola_2017_image}
Isola P, Zhu J-Y, Zhou T, et~al.
\newblock Image-to-Image Translation with Conditional Adversarial Networks.
\newblock In: Proceedings of the IEEE Conference on Computer Vision and Pattern
  Recognition; 2017. p. 5967--5976.

\bibitem{zhu_unpair_2017}
Zhu J-Y, Park T, Isola P, et~al.
\newblock Unpaired Image-to-Image Translation Using Cycle-Consistent
  Adversarial Networks.
\newblock In: Proceedings of the IEEE International Conference on Computer
  Vision; 2017. p. 2242--2251.

\bibitem{Nakamura_2017_scene}
Nakamura T, Zhu A, Yanai K, et~al.
\newblock Scene Text Eraser.
\newblock In: Proceedings of the International Conference on Document Analysis
  and Recognition. vol.~01; 2017. p. 832--837.

\bibitem{Zhang_2019_ensnet}
Zhang S, Liu Y, Jin L, et~al.
\newblock EnsNet: Ensconce Text in the Wild.
\newblock In: Proceedings of the AAAI Conference on Artificial Intelligence;
  2019. p. 801--808.

\bibitem{Tursun_2019_mtrnet}
Tursun O, Zeng R, Denman S, et~al.
\newblock MTRNet: A Generic Scene Text Eraser.
\newblock In: Proceedings of the International Conference on Document Analysis
  and Recognition; 2019. .

\bibitem{Iizuka_2017_global}
Iizuka S, Simo-Serra E, Ishikawa H.
\newblock Globally and Locally Consistent Image Completion.
\newblock ACM Transactions on Graphics. 2017;36(4):107:1--107:14.

\bibitem{Yu_2018_CVPR}
Yu J, Lin Z, Yang J, et~al.
\newblock Generative Image Inpainting With Contextual Attention.
\newblock In: Proceedings of the IEEE Conference on Computer Vision and Pattern
  Recognition; 2018. .

\bibitem{ye_text_2015}
Ye Q, Doermann D.
\newblock Text Detection and Recognition in Imagery: A Survey.
\newblock IEEE Transactions on Pattern Analysis and Machine Intelligence.
  2015;37(7):1480--1500.

\bibitem{shi2017detect}
Shi B, Bai X, Belongie S.
\newblock Detecting Oriented Text in Natural Images by Linking Segments.
\newblock In: Proceedings of the IEEE Conference on Computer Vision and Pattern
  Recognition; 2017. p. 3482--3490.

\bibitem{liu_curved_2019}
Liu Y, Jin L, Zhang S, et~al.
\newblock Curved scene text detection via transverse and longitudinal sequence
  connection.
\newblock Pattern Recognition. 2019;90:337 -- 345.

\bibitem{chen2019irregular}
Chen J, Lian Z, Wang Y, et~al.
\newblock Irregular scene text detection via attention guided border labeling.
\newblock Science China Information Sciences. 2019;62(12):220103--.

\bibitem{he_realtime_2020}
He W, Zhang X-Y, Yin F, et~al.
\newblock Realtime multi-scale scene text detection with scale-based region
  proposal network.
\newblock Pattern Recognition. 2020;98:107026.

\bibitem{baek_2019_craft}
Baek Y, Lee B, Han D, et~al.
\newblock Character Region Awareness for Text Detection.
\newblock In: Proceedings of the IEEE Conference on Computer Vision and Pattern
  Recognition; 2019. p. 9365--9374.

\bibitem{zhang_automatic_2015}
Zhang C, Yao C, Shi B, et~al.
\newblock Automatic discrimination of text and non-text natural images.
\newblock In: Proceedings of the International Conference on Document Analysis
  and Recognition; 2015. p. 886--890.

\bibitem{mser2004}
Matas J, Chum O, Urban M, et~al.
\newblock Robust wide-baseline stereo from maximally stable extremal regions.
\newblock Image and Vision Computing. 2004;22(10):761 -- 767.

\bibitem{bow1998}
Joachims T.
\newblock Text categorization with Support Vector Machines: Learning with many
  relevant features.
\newblock In: European Conference on Machine Learning; 1998. p. 137--142.

\bibitem{bai_text_2017}
Bai X, Shi B, Zhang C, et~al.
\newblock Text/non-text image classification in the wild with convolutional
  neural networks.
\newblock Pattern Recognition. 2017;66:437 -- 446.

\bibitem{zhao_fast_2019}
Zhao M, Wang R-Q, Yin F, et~al.
\newblock Fast Text/non-Text Image Classification with Knowledge Distillation.
\newblock In: Proceedings of the International Conference on Document Analysis
  and Recognition; 2019. p. 1458--1463.

\bibitem{gupta_text_2020}
Gupta N, Jalal A~S.
\newblock Text or Non-text Image Classification using Fully Convolution Network
  (FCN).
\newblock In: Proceedings of the International Conference on Contemporary
  Computing and Applications; 2020. p. 150--153.

\bibitem{Zhou_2017_EAST}
Zhou X, Yao C, Wen H, et~al.
\newblock EAST: An Efficient and Accurate Scene Text Detector.
\newblock In: Proceedings of the IEEE Conference on Computer Vision and Pattern
  Recognition; 2017. p. 2642--2651.

\bibitem{Ren_2019_ICCV}
Ren Y, Yu X, Zhang R, et~al.
\newblock StructureFlow: Image Inpainting via Structure-Aware Appearance Flow.
\newblock In: Proceedings of the IEEE International Conference on Computer
  Vision; 2019. .

\bibitem{Yu_2018_freeform}
Yu J, Lin Z, Yang J, et~al.
\newblock Free-Form Image Inpainting With Gated Convolution.
\newblock In: Proceedings of the IEEE International Conference on Computer
  Vision; 2019. .

\bibitem{unet}
Ronneberger O, Fischer P, Brox T.
\newblock U-Net: Convolutional Networks for Biomedical Image Segmentation.
\newblock In: Medical Image Computing and Computer-Assisted Intervention; 2015.
  p. 234--241.

\bibitem{Miyato-2018-spectral}
Miyato T, Kataoka T, Koyama M, et~al.
\newblock Spectral Normalization for Generative Adversarial Networks.
\newblock In: International Conference on Learning Representations; 2018. .

\bibitem{Tran_2017_hier}
Tran D, Ranganath R, Blei D.
\newblock Hierarchical Implicit Models and Likelihood-Free Variational
  Inference.
\newblock In: Proceedings of the Advances in Neural Information Processing
  Systems; 2017. p. 5523--5533.

\bibitem{Zhang_2019_self}
Zhang H, Goodfellow I, Metaxas D, et~al.
\newblock Self-Attention Generative Adversarial Networks.
\newblock In: Proceedings of the International Conference on Machine Learning;
  2019. p. 7354--7363.

\bibitem{Gatys_2016_image}
Gatys L~A, Ecker A~S, Bethge M.
\newblock Image Style Transfer Using Convolutional Neural Networks.
\newblock In: Proceedings of the IEEE Conference on Computer Vision and Pattern
  Recognition; 2016. p. 2414--2423.

\bibitem{Aly_2005_image}
Aly H~A, Dubois E.
\newblock Image up-sampling using total-variation regularization with a new
  observation model.
\newblock IEEE Transactions on Image Processing. 2005;14(10):1647--1659.

\bibitem{Gupta_2016_synthetic}
Gupta A, Vedaldi A, Zisserman A.
\newblock Synthetic Data for Text Localisation in Natural Images.
\newblock In: Proceedings of the IEEE Conference on Computer Vision and Pattern
  Recognition; 2016. p. 2315--2324.

\bibitem{Nayef_2017_mlt}
Nayef N, Yin F, Bizid I, et~al.
\newblock ICDAR2017 Robust Reading Challenge on Multi-Lingual Scene Text
  Detection and Script Identification - RRC-MLT.
\newblock In: Proceedings of the IAPR International Conference on Document
  Analysis and Recognition; 2017. p. 1454--1459.

\bibitem{dutta2019via}
Dutta A, Zisserman A.
\newblock The VIA annotation software for images, audio and video.
\newblock In: Proceedings of the 27th ACM International Conference on
  Multimedia; 2019. p. 2276--2279.

\bibitem{Wolf_2006_image}
Wolf C, Jolion J-M.
\newblock Object count/Area Graphs for the Evaluation of Object Detection and
  Segmentation Algorithms.
\newblock International Journal of Document Analysis and Recognition.
  2006;8(4):280–296.

\bibitem{Kingma_2015_adam}
Kingma D~P, Ba J.
\newblock Adam: A method for stochastic optimization.
\newblock In: International Conference on Learning Representations; 2015. .

\end{thebibliography}

\end{document}